\pdfoutput=1

\documentclass[11pt]{article}

\usepackage[]{emnlp2021}

\usepackage{times}
\usepackage{latexsym}

\usepackage[T1]{fontenc}

\usepackage[utf8]{inputenc}

\usepackage{microtype}

\usepackage{graphicx}
\usepackage{booktabs,fancyvrb}
\usepackage{multirow}
\usepackage{inconsolata}
\usepackage{xspace}
\usepackage{subcaption}
\usepackage{amsmath}
\usepackage{cleveref}
\usepackage{paralist}
\usepackage{enumitem}
\usepackage[normalem]{ulem}
\usepackage{xparse}
\usepackage{makecell}
\NewDocumentCommand\ulverb{v}{\uline{\ttfamily#1}}

\title{Exploring Underexplored Limitations of Cross-Domain Text-to-SQL Generalization}

\author{Yujian Gan${}^{1}$ \ \ \ \ Xinyun Chen${}^{2}$ \ \ \ \  Matthew Purver${}^{1,3}$   \\
${}^{1}$Queen Mary University of London \ \ \ \  \ \ \ \ ${}^{2}$UC Berkeley
\ \ \  \ \ \ \ ${}^{3}$Jožef Stefan Institute\\
  \texttt{\{y.gan,m.purver\}@qmul.ac.uk}  \ \ \ \  \ \ \ \  \texttt{xinyun.chen@berkeley.edu} \\
  }

\begin{document}
\maketitle
\begin{abstract}
Recently, there has been significant progress in studying neural networks for translating text descriptions into SQL queries under the zero-shot cross-domain setting.
Despite achieving good performance on some public benchmarks, we observe that existing text-to-SQL models do not generalize when facing domain knowledge that does not frequently appear in the training data, which may render the worse prediction performance for unseen domains.
In this work, we investigate the robustness of text-to-SQL models when the questions require rarely observed domain knowledge.
In particular, we define five types of domain knowledge and introduce Spider-DK (DK is the abbreviation of domain knowledge), a human-curated dataset based on the Spider benchmark for text-to-SQL translation. 
NL questions in Spider-DK are selected from Spider, and we modify some samples by adding domain knowledge that reflects real-world question paraphrases. 
We demonstrate that the prediction accuracy dramatically drops on samples that require such domain knowledge, even if the domain knowledge appears in the training set, and the model provides the correct predictions for related training samples.
\footnote{Our dataset is available at  \href{https://github.com/ygan/Spider-DK}{https://github.com/ygan/Spider-DK}.}
\end{abstract}

\section{Introduction}

Research on cross-domain text-to-SQL benchmarks has led to numerous advances. 
Recent works \cite{DBLP:journals/corr/abs-2101-09901,DBLP:journals/corr/abs-2010-12412,DBLP:journals/corr/abs-2103-04399} have achieved over 70\% accuracy on Spider benchmark~\cite{Yu2018a} and over 90\% accuracy on WikiSQL benchmark~\cite{zhongSeq2SQL2017}, which seems to suggest that existing models already solved most problems in this field.
However, the follow-up studies from \citet{Deng2020,gan-etal-2021-towards,Suhr2020,Shaw2020,oren-etal-2020-improving,keysers2020measuring} show that the generalization performance is much worse in more challenging scenarios.
For example, \citet{Deng2020} investigate the cases when the explicit mentions of database columns are removed from the question.
Similarly, \cite{gan-etal-2021-towards} observe that the model accuracy dramatically drops by replacing schema-related words with some synonyms.
On the other hand, \citet{Suhr2020} find that the generalization to other databases is much worse, due to the distribution shift of both questions and SQL queries.
These papers introduce important challenges for improving the generalization performance, i.e., the model trained on a cross-domain text-to-SQL dataset (e.g., Spider \cite{Yu2018a}) does not generalize to a new external database.
However, the performance degradation is somehow expected for the following reasons. First, removing the explicit mentions breaks the assumptions that make the schema linking effective. Second, SQL queries in other databases could come from a different distribution; e.g., according to the hardness criteria defined by Spider benchmark, over 40\% Spider SQL queries are \verb|Medium| hardness, but there are less than 10\% \verb|Medium| SQL queries in the GeoQuery dataset \cite{data-geography-original}.

In this work, we demonstrate that the generalization performance could be poor even when both the NL questions and SQL queries follow the similar distribution to the training set. 
Specifically, we constructed Spider-DK, a challenging variant of the Spider development set, with the focus of evaluating the model understanding of domain knowledge. 
A domain means a certain type of application scenarios; for example, the Spider benchmark includes various distinct domains such as geography and university. Cross-domain text-to-SQL research aims to build a text-to-SQL model that can generate correct SQL queries and generalize to different domains. Therefore, one main challenge of cross-domain text-to-SQL generalization is to understand different knowledge required by different domains. For example, the university domain usually needs the knowledge of different job titles and genders, while the geography domain emphasizes more on the knowledge of places instead of people.
We show that the state-of-the-art models consistently fail in cases when specific domain knowledge is required for prediction, even if the domain knowledge is moderately mentioned in the training data, and the models accurately predict the corresponding training samples. Such discrepancy suggests that the models do not properly learn the domain knowledge in order to fit the training set, thus improving the model capability to capture the domain knowledge is an important direction towards achieving the cross-domain generalization for text-to-SQL applications.
To our knowledge, we are the first work investigating the text-to-SQL model capability of understanding the domain knowledge provided in the training set, and generalizing the knowledge to new problems.

\begin{table}[t]
    \centering
   
    \resizebox{.99\columnwidth}{!}{
    \smallskip\begin{tabular}{cl}
        \\ \specialrule{0.08em}{0pt}{4pt} 
      {\bf T1} & \makecell[c]{SELECT Columns Mentioned by Omission}
       \\
      {\bf NL} & \textit{Find the \uline{name}  of the teacher who \textbf{\textrm{...}}}  \\ 
      {\bf SQL} & \ttfamily select  \ulverb|firstname| , \ulverb|lastname| from \textbf{\textrm{...}} 
      \\ \specialrule{0.05em}{4pt}{4pt} 

      {\bf T2} & \makecell[c]{Simple Inference Required}
       \\
      {\bf NL} & \textbf{\textrm{...}} \textit{order of their date of birth \uline{from old to young}.}  \\
      {\bf SQL} & \textbf{\textrm{...}}  \ttfamily order by date\_of\_birth \ulverb|asc|   
      \\ \specialrule{0.05em}{4pt}{4pt} 

      {\bf T3} & 
      \makecell[c]{Synonyms Substitution in Cell Value Word}\\
      {\bf NL} &  \textit{List the state in the \uline{US}} \textbf{\textrm{...}} \\
      {\bf SQL} & \textbf{\textrm{...}}  \ttfamily where billing\_country  =   \ulverb|"USA"| \textbf{\textrm{...}}  
      \\ \specialrule{0.05em}{4pt}{4pt} 

      {\bf T4} & 
      \makecell[c]{One Non-Cell Value Word Generate a Condition} \\
      {\bf NL} & \textit{How many students got \uline{accepted} after the tryout?}  \\
      {\bf SQL} &  \textbf{\textrm{...}} \ttfamily  from tryout where \ulverb|decision="yes"|    
      \\ \specialrule{0.05em}{4pt}{4pt} 

      {\bf T5} & 
      \makecell[c]{Easy to Conflict with other Domains}  \\
      {\bf NL} & \textbf{\textrm{...}} \textit{with \uline{max speed}  higher than 1000.}  \\
      {\bf SQL} & \textbf{\textrm{...}}  \ttfamily where \ulverb|max_speed|  >  1000    \\     
      
      \bottomrule 
    \end{tabular}
    }
    \caption{Five types of domain knowledge extracted from Spider training set. We name them as T1 to T5.}
  
    \label{table:example-of-dk}
\end{table}

\section{Spider-DK Dataset}

\subsection{Overview}
\label{section:2Overall}

We construct the Spider-DK benchmark by selecting samples from the Spider development set that require domain knowledge understanding, and we also manually modify some samples to incorporate domain knowledge.
The purpose of building Spider-DK is to simulate the scenario where specific domain knowledge is involved in the users' utterance query.
Domain knowledge is often used unnoticedly, which makes some domain knowledge unavoidable.
For example, in the T5 of Table \ref{table:example-of-dk}, the direct use of the \verb|max_speed| column annotation raises a domain knowledge problem.
We discuss the details of this problem later in Section \ref{sec:dk}.

Spider-DK contains 535 NL-SQL pairs drawn from the Spider development set, where 270 pairs are the same as the original Spider samples, while the rest 265 pairs are modified to incorporate the domain knowledge.
We categorize the types of domain knowledge required in Spider-DK, which makes it easy for breakdown analysis.
Spider-DK is smaller than the Spider development set, because not every domain or example can be easily modified to incorporate some domain knowledge. 
Besides, it is hard to evaluate the model generalization ability for domain knowledge if keeping too many original Spider examples that do not require domain knowledge.

In particular, the distribution of the SQL query hardness in Spider-DK is close to the original Spider, i.e., \verb|easy| accounts for 20.6\%, \verb|medium| accounts for 41.8\%, \verb|hard| accounts for 14.8\%, and \verb|extra hard| accounts for 19.1\%~\footnote{The Spider benchmark defines four hardness levels.}.  
We define five types of domain knowledge in Table \ref{table:example-of-dk}. In Spider-DK, \verb|T1| accounts for 28.7\% of samples, \verb|T2| accounts for 24.5\%, \verb|T3| accounts for 27.5\%, \verb|T4| accounts for 8.3\%, and \verb|T5| accounts for 12.5\%.

We curate the Spider-DK by modifying only questions or both questions and SQL queries, as shown in Table \ref{table:spider2spiderdk}.
We carefully add the domain knowledge into the utterance to ensure that the new utterance follows the domain knowledge required by existing Spider samples and does not raise ambiguity.
Most domain knowledge in Spider-DK is similar to that in the Spider training set. 
Compared to the evaluation sets in \cite{Suhr2020}, Spider-DK is easier and closer to the training data and focuses only on domain knowledge, and we provide more discussion below.

\begin{table}[t]
    \centering
   
    \resizebox{.99\columnwidth}{!}{
    \smallskip\begin{tabular}{cl}
        \\ \specialrule{0.08em}{0pt}{4pt} 
        \multicolumn{2}{c}{Only Modify the NL } 
       \\
      {\bf Spider} & \textbf{\textrm{...}} \textit{in the order of birth date.}  \\ 
      {\bf Spider-DK} & \textbf{\textrm{...}} \textit{order of their birth date from old to young.}  
      \\ \specialrule{0.05em}{4pt}{4pt} 

        \multicolumn{2}{c}{Modify both NL and SQL } 
       \\
      {\bf Spider} & \textit{Compute the average age of dogs.}  \\ 
      & \ttfamily  select avg(age) from dogs \\
      {\bf Spider-DK} & \textit{Compute the average age of \uline{abandoned} dogs.}      \\     
      & \ttfamily  select avg(age) from dogs \\
      & \ttfamily  where {\uline{\ttfamily abandoned\_y = 1}} \\

      \bottomrule 
    \end{tabular}
    }
    \caption{Examples of Spider question and/or SQL modifications made in Spider-DK.}
  
    \label{table:spider2spiderdk}
\end{table}

\subsection{Domain Knowledge}
\label{sec:dk}
Different SQL databases could require very different domain knowledge. As shown in~\cite{Suhr2020}, the state-of-the-art models on Spider achieve much worse performance on earlier SQL benchmarks such as ATIS and GeoQuery~\cite{data-atis-geography-scholar,data-geography-original}. However, we argue that the failure of generalization is expected to some extent, because without seeing in-domain examples, some domain knowledge required by these datasets is even hard to infer for experienced programmers.
For example, we asked five computer science graduate students to write the SQL query for the question \verb|`how many major cities are there?'| in GeoQuery, but none of them gave the correct answer.
This question requires the domain knowledge that \verb|major| means \verb|`population > 150000'|, which is hard to infer without looking at GeoQuery training set.
Therefore, while acquiring general-purpose domain knowledge is also important, we believe that the failure of generalization to questions requiring similar domain knowledge to the training set could be more problematic, which motivates our design of Spider-DK benchmark.

We study five types of domain knowledge (name them as T1 to T5) shown in Table \ref{table:example-of-dk}. T1 requires the models to understand that the user queries two columns by an omitted expression. 

T2 requires the models to infer the correct queries, e.g., if the T2 utterance in Table \ref{table:example-of-dk} modified from \verb|`date of birth'| to \verb|`age'|, the model should output \verb|desc| not \verb|asc|. 
Note that the Spider training set contains both \verb|`date of birth'| and \verb|`age'| along with \verb|`old to young'|.

T3 requires the models to recognize the cell value synonym substitution. 
Some synonym substitutions base on their adjective form, such as \verb|`singer whose country is France'| and \verb|`French singer'|. 

Although the number of T4 is the least in Spider-DK, it is not uncommon in the Spider training set. 
Unlike the GeoQuery \verb|major| example mentioned above, T4 only includes the conditions whose column type is similar to boolean. 
For example, in Table \ref{table:example-of-dk} and \ref{table:spider2spiderdk}, the column \verb|decision| only contain \verb|yes| and \verb|no|, while \verb|abandoned_y| only contain \verb|1| and \verb|0|.
Therefore, the key to solving T4 is whether the model can distinguish whether the column is a boolean-like type, but the difficulty is that the word varies in different domains.

Although T5 seems simple and does not seem to contain domain knowledge, the models that generate SQL structure and schema items separately are easy to mispredict in T5. 
A review \cite{gan-etal-2020-review} shows that most models follow the separate generation pattern, i.e., these models may use the same word twice in both generating schema items and SQL structure.
Because, in other domain training data, the models learn to generate a \verb|max()| function when the utterance contains a word \verb|max|.
Therefore, these models may use the word \verb|max| twice to generate the \verb|max(max_speed)| for T5 utterance instead of a simple \verb|max_speed|.

\section{Experiments}
\subsection{Experimental Setup}

We evaluate the previous state-of-the-art models on the Spider-DK and Spider \cite{Yu2018a}.
As discussed in Section~\ref{section:2Overall}, the Spider test set is not publicly accessible, and thus Spider-DK does not contain a test set.
We extracted 535 examples corresponding to Spider-DK from Spider for evaluation instead of using a whole Spider development set for better comparison. 
In addition, we select 125 examples with domain knowledge from the training set to evaluate the training effect.
Therefore, there are three evaluation sets: 
\begin{itemize}[leftmargin=*,noitemsep,topsep=0em]
    \item $ \textbf{Spider}_{\textnormal{\small{T}}}$: 125 examples drawn from the Spider training set.  
    \item $ \textbf{Spider}_{\textnormal{\small{D}}}$:  535 examples drawn from the Spider development set.
    \item {\bf Spider-DK}: Spider-DK development set with 535 examples. 
\end{itemize}

We evaluate open-source models that reach competitive performance on Spider: GNN \cite{Bogin2019}, IRNet \cite{Guo2019}, RAT-SQL \cite{Wang2019} with and without BERT \cite{Kenton2017}, and RAT-SQL + GAP \cite{DBLP:journals/corr/abs-2012-10309}.
We present their results of the 265 Spider-DK domain knowledge examples and analyze their performance in each knowledge type.
Our evaluation is based on the exact match metric defined in the original Spider benchmark, which measures whether the predicted query without condition values as a whole is equivalent to the gold query.

\begin{table}[t]
    \centering
    \resizebox{.99\columnwidth}{!}{
    \smallskip\begin{tabular}{lccc}
        \hline
       \bf model & \bf $ \textbf{Spider}_{\textnormal{\small{T}}}$ &\bf $ \textbf{Spider}_{\textnormal{\small{D}}}$ & \bf Spider-DK \\
        \hline \hline
        GNN  \cite{Bogin2019}  & 61.6\% & 46.2\% & 26.0\% \\ 
        IRNet  \cite{Guo2019}   & 87.2\% & 53.8\% & 33.1\% \\ 
        RAT-SQL  \cite{Wang2019}  &  93.6\% & 61.1\% &  35.8\% \\ 
        RAT-SQL + BERT \cite{Wang2019} &  92.0\% & \bf 73.3\% &  40.9\% \\ 
        RAT-SQL  + GAP \cite{DBLP:journals/corr/abs-2012-10309} & \bf 98.4\% &  67.8\% & \bf 44.1\% \\ 
        \hline 
    \end{tabular}
    }
    \caption{Exact match accuracy on the $ \textbf{Spider}_{\textnormal{\small{T}}}$, $ \textbf{Spider}_{\textnormal{\small{D}}}$ and {\bf Spider-DK}, where models are trained on the original Spider training set.}\smallskip
    \label{table:previous-models}
    
\end{table}

\subsection{Main Results}
Table \ref{table:previous-models} presents the exact match accuracy of different models on $ \textnormal{Spider}_{\textnormal{\small{T}}}$, $ \textnormal{Spider}_{\textnormal{\small{D}}}$, and Spider-DK. All models are trained on the original Spider training set. 
Compared to $ \textnormal{Spider}_{\textnormal{\small{D}}}$, the performance of all models has significantly dropped by about 20\% to 30\% on Spider-DK.
Although the Spider-DK is designed based on the $ \textnormal{Spider}_{\textnormal{\small{T}}}$, whose  exact match evaluation is pretty high, these models can not generalize to the Spider-DK well.

In particular, although RAT-SQL + BERT achieves better performance on $ \textnormal{Spider}_{\textnormal{\small{D}}}$ than RAT-SQL + GAP, RAT-SQL + GAP outperforms RAT-SQL + BERT on Spider-DK, indicating that GAP facilitates the model to grasp a better understanding of domain knowledge.
Despite some improvement achieved by recent models, the results show that domain knowledge understanding is still a considerable gap toward the realization of cross-domain text-to-SQL generation.

\begin{table}[t]
    \centering
    \resizebox{.99\columnwidth}{!}{
    \smallskip\begin{tabular}{lcccccc}
        \hline
       \bf Approach & \bf ALL & \bf T1 & \bf T2 & \bf T3 & \bf T4 & \bf T5 \\
        \hline \hline
        GNN  &  6.8\%&  5.3\% &  7.6\% & 2.7\% &  8.3\% & 21.2\%\\ 
        IRNet  &  19.2\%& \bf 9.2\% &   4.6\% & 42.4\% & 4.5\% & 27.2\%\\ 
        RAT-SQL &  16.6\%&  2.6\% &  13.8\% &   26.0\% &  \bf 9.1\% & 36.4\%\\ 
        RAT-SQL + BERT &  19.6\% &  3.9\%  &  12.3\% & 41.1\% & 4.5\% & 30.3\%\\ 
        RAT-SQL + GAP & \bf 27.1\% &  7.9\%  & \bf 20.0\% & \bf 53.4\% & \bf 9.1\% &  \bf42.4\%\\ 
        
        \hline 
    \end{tabular}
    }
    \caption{Break down exact match accuracy in the Spider-DK examples containing domain knowledge.}\smallskip
    \label{table:break-down-results}
\end{table}

\subsection{Performance on Knowledge Type Splits}
To better understand the performance facing the domain knowledge, we present the breakdown accuracies of different domain knowledge types in Table \ref{table:break-down-results}.
RAT-SQL + GAP unsurprisingly achieves the best performance on all examples and outperforms other models from T2 to T5.
However, IRNet surprisingly obtains an overall accuracy close to the RAT-SQL + BERT, because IRNet integrates a ConceptNet~\cite{speer-havasi-2012-representing} to recognize the country, state, and city synonyms, which can improve its accuracy in T3.
The GNN and RAT-SQL perform relatively poorly on T3 because they do not have extra knowledge components such as ConceptNet. 
Besides, GNN trains its embeddings from scratch, and RAT-SQL uses GLOVE~\cite{pennington-etal-2014-glove} that has been shown worse than BERT in many scenarios.
Although ConceptNet helps IRNet in T3, it is not a generalization method for solving other domain knowledge problems.
However, even the best-performing T3 is still far from the accuracy in $ \textnormal{Spider}_{\textnormal{\small{D}}}$, which shows that there is still much room for improvement.

\subsection{Error Analysis}

Table \ref{table:example-error} presents five error examples in each knowledge type drawn from the prediction of RAT-SQL + GAP. 
These error predictions are similar to the training examples shown in Table \ref{table:example-of-dk}.
There are three reasons why existing models can not perform well in the Spider-DK.
The first reason is that some domain knowledge is not common enough in the training set. 
For example, in T2, the phrase \verb|`from old to young'| appears more often with \verb|age|, which trains the model to output a \verb|desc age| order.
The unbalance training data may lead the model to prefer outputting a \verb|desc| order even its column is the \verb|`date of birth'|. 

The second reason is that the model has insufficient generalization ability for similar problems. Many training examples belong to the T3 and T4. However, these examples can not cover all cases. 
For example, the training data may not or rarely contain examples where the USA is substituted with the United States, but we expect the models can still handle these examples correctly. 
The third reason is that a word will be used twice to generate schema items and SQL structure as we discussed the T5 in Section \ref{sec:dk}.

\begin{table}[t]
    \centering
   
    \resizebox{.99\columnwidth}{!}{
    \smallskip\begin{tabular}{cl}
        \\ \specialrule{0.08em}{0pt}{4pt} 
      {\bf (T1)NL} & \textit{Find the \uline{name} of the professionals \textbf{\textrm{...}}}  \\ 
      {\bf Pred} & \ttfamily select  \ulverb|first_name| from \textbf{\textrm{...}} \\ 
      {\bf Gold} & \ttfamily select  \ulverb|first_name| , \ulverb|last_name| from \textbf{\textrm{...}} 
      \\ \specialrule{0.05em}{4pt}{4pt} 

      {\bf (T2)NL} & \textbf{\textrm{...}} \textit{sorted  \uline{from oldest to youngest}? }  \\ 
      {\bf Pred} & \textbf{\textrm{...}}  \ttfamily order by birth\_date \ulverb|desc| \\
      {\bf Gold} & \textbf{\textrm{...}}  \ttfamily order by birth\_date \ulverb|asc| 
      \\ \specialrule{0.05em}{4pt}{4pt} 

      {\bf (T3)NL} & \textit{List all \uline{American} airline names and their abbreviations.}  \\ 
      {\bf Pred} & \ttfamily select airline, abbreviation from airlines \\ 
      {\bf Gold} & \ttfamily select airline, abbreviation from airlines \\ 
      & \ttfamily where  {\uline{\ttfamily country = `USA'}} 
      \\ \specialrule{0.05em}{4pt}{4pt} 

      {\bf (T4)NL} & \textit{What is the average age of all the \uline{abandoned} dogs?}  \\ 
      {\bf Pred} & \ttfamily select avg(age) from dogs \\ 
      {\bf Gold} & \ttfamily select avg(age) from dogs \\ & \ttfamily  where {\uline{\ttfamily abandoned\_y = 1}} 
      \\ \specialrule{0.05em}{4pt}{4pt} 

      {\bf (T5)NL} & \textit{Show \uline{average} attendance for all stadiums \textbf{\textrm{...}}}  \\ 
      {\bf Pred} & \ttfamily select  \ulverb|avg(average)|  from stadium \textbf{\textrm{...}} \\ 
      {\bf Gold} & \ttfamily select  \ulverb|average|  from stadium \textbf{\textrm{...}} \\
      \bottomrule 
    \end{tabular}
    }
    \caption{Sample wrong predictions of RAT-SQL + GAP in each type of domain knowledge.}
  
    \label{table:example-error}
\end{table}

\section{Conclusion}
We introduce Spider-DK, a human-curated dataset based on the Spider benchmark for evaluating the generalization of text-to-SQL models, with the focus of understanding the domain knowledge. 
We demonstrate that the performance of existing text-to-SQL models drops dramatically on Spider-DK, even if the domain knowledge appears in the training set.
Our evaluation indicates that the models do not always understand the underlying domain knowledge for prediction, thus we consider improving the model understanding of domain knowledge as an important direction to achieve the cross-domain generalization.

\section*{Acknowledgements}
We would like to thank the anonymous reviewers for their helpful comments.
Matthew Purver is partially supported by the EPSRC under grant EP/S033564/1, and by the European Union's Horizon 2020 programme under grant agreement 
825153 (EMBEDDIA, Cross-Lingual Embeddings for Less-Represented Languages in European News Media). Xinyun Chen is supported by the Facebook Fellowship. The results of this publication reflect only the authors' views and the Commission is not responsible for any use that may be made of the information it contains.

\bibliography{emnlp2021}

\begin{thebibliography}{21}
\expandafter\ifx\csname natexlab\endcsname\relax\def\natexlab#1{#1}\fi

\bibitem[{Bogin et~al.(2019)Bogin, Berant, and Gardner}]{Bogin2019}
Ben Bogin, Jonathan Berant, and Matt Gardner. 2019.
\newblock \href {https://doi.org/10.18653/v1/P19-1448} {Representing schema
  structure with graph neural networks for text-to-{SQL} parsing}.
\newblock In \emph{Proceedings of the 57th Annual Meeting of the Association
  for Computational Linguistics}, pages 4560--4565, Florence, Italy.
  Association for Computational Linguistics.

\bibitem[{Deng et~al.(2021)Deng, Awadallah, Meek, Polozov, Sun, and
  Richardson}]{Deng2020}
Xiang Deng, Ahmed~Hassan Awadallah, Christopher Meek, Oleksandr Polozov, Huan
  Sun, and Matthew Richardson. 2021.
\newblock \href {https://doi.org/10.18653/v1/2021.naacl-main.105}
  {Structure-grounded pretraining for text-to-{SQL}}.
\newblock In \emph{Proceedings of the 2021 Conference of the North American
  Chapter of the Association for Computational Linguistics: Human Language
  Technologies}, pages 1337--1350, Online. Association for Computational
  Linguistics.

\bibitem[{Devlin et~al.(2019)Devlin, Chang, Lee, and Toutanova}]{Kenton2017}
Jacob Devlin, Ming-Wei Chang, Kenton Lee, and Kristina Toutanova. 2019.
\newblock \href {https://doi.org/10.18653/v1/N19-1423} {{BERT}: Pre-training of
  deep bidirectional transformers for language understanding}.
\newblock In \emph{Proceedings of the 2019 Conference of the North {A}merican
  Chapter of the Association for Computational Linguistics: Human Language
  Technologies, Volume 1 (Long and Short Papers)}, pages 4171--4186,
  Minneapolis, Minnesota. Association for Computational Linguistics.

\bibitem[{Gan et~al.(2021)Gan, Chen, Huang, Purver, Woodward, Xie, and
  Huang}]{gan-etal-2021-towards}
Yujian Gan, Xinyun Chen, Qiuping Huang, Matthew Purver, John~R. Woodward,
  Jinxia Xie, and Pengsheng Huang. 2021.
\newblock \href {https://doi.org/10.18653/v1/2021.acl-long.195} {Towards
  robustness of text-to-{SQL} models against synonym substitution}.
\newblock In \emph{Proceedings of the 59th Annual Meeting of the Association
  for Computational Linguistics and the 11th International Joint Conference on
  Natural Language Processing (Volume 1: Long Papers)}, pages 2505--2515,
  Online. Association for Computational Linguistics.

\bibitem[{Gan et~al.(2020)Gan, Purver, and Woodward}]{gan-etal-2020-review}
Yujian Gan, Matthew Purver, and John~R. Woodward. 2020.
\newblock \href {https://www.aclweb.org/anthology/2020.aacl-srw.16} {A review
  of cross-domain text-to-{SQL} models}.
\newblock In \emph{Proceedings of the 1st Conference of the Asia-Pacific
  Chapter of the Association for Computational Linguistics and the 10th
  International Joint Conference on Natural Language Processing: Student
  Research Workshop}, pages 108--115, Suzhou, China. Association for
  Computational Linguistics.

\bibitem[{Guo et~al.(2019)Guo, Zhan, Gao, Xiao, Lou, Liu, and Zhang}]{Guo2019}
Jiaqi Guo, Zecheng Zhan, Yan Gao, Yan Xiao, Jian-Guang Lou, Ting Liu, and
  Dongmei Zhang. 2019.
\newblock \href {https://doi.org/10.18653/v1/P19-1444} {{Towards Complex
  Text-to-{SQL} in Cross-Domain Database with Intermediate Representation}}.
\newblock In \emph{Proceedings of the 57th Annual Meeting of the Association
  for Computational Linguistics}, pages 4524--4535, Florence, Italy.
  Association for Computational Linguistics.

\bibitem[{Hui et~al.(2021)Hui, Shi, Geng, Li, Li, Sun, and
  Zhu}]{DBLP:journals/corr/abs-2103-04399}
Binyuan Hui, Xiang Shi, Ruiying Geng, Binhua Li, Yongbin Li, Jian Sun, and
  Xiaodan Zhu. 2021.
\newblock \href {http://arxiv.org/abs/2103.04399} {Improving text-to-sql with
  schema dependency learning}.
\newblock \emph{CoRR}, abs/2103.04399.

\bibitem[{Iyer et~al.(2017)Iyer, Konstas, Cheung, Krishnamurthy, and
  Zettlemoyer}]{data-atis-geography-scholar}
Srinivasan Iyer, Ioannis Konstas, Alvin Cheung, Jayant Krishnamurthy, and Luke
  Zettlemoyer. 2017.
\newblock \href {https://doi.org/10.18653/v1/P17-1089} {{Learning a Neural
  Semantic Parser from User Feedback}}.
\newblock In \emph{Proceedings of the 55th Annual Meeting of the Association
  for Computational Linguistics (Volume 1: Long Papers)}, pages 963--973.

\bibitem[{Keysers et~al.(2020)Keysers, Sch{\"a}rli, Scales, Buisman, Furrer,
  Kashubin, Momchev, Sinopalnikov, Stafiniak, Tihon, Tsarkov, Wang, van Zee,
  and Bousquet}]{keysers2020measuring}
Daniel Keysers, Nathanael Sch{\"a}rli, Nathan Scales, Hylke Buisman, Daniel
  Furrer, Sergii Kashubin, Nikola Momchev, Danila Sinopalnikov, Lukasz
  Stafiniak, Tibor Tihon, Dmitry Tsarkov, Xiao Wang, Marc van Zee, and Olivier
  Bousquet. 2020.
\newblock \href {https://openreview.net/forum?id=SygcCnNKwr} {Measuring
  compositional generalization: A comprehensive method on realistic data}.
\newblock In \emph{International Conference on Learning Representations}.

\bibitem[{Oren et~al.(2020)Oren, Herzig, Gupta, Gardner, and
  Berant}]{oren-etal-2020-improving}
Inbar Oren, Jonathan Herzig, Nitish Gupta, Matt Gardner, and Jonathan Berant.
  2020.
\newblock \href {https://doi.org/10.18653/v1/2020.findings-emnlp.225}
  {Improving compositional generalization in semantic parsing}.
\newblock In \emph{Findings of the Association for Computational Linguistics:
  EMNLP 2020}, pages 2482--2495, Online. Association for Computational
  Linguistics.

\bibitem[{Pennington et~al.(2014)Pennington, Socher, and
  Manning}]{pennington-etal-2014-glove}
Jeffrey Pennington, Richard Socher, and Christopher Manning. 2014.
\newblock \href {https://doi.org/10.3115/v1/D14-1162} {{G}love: Global vectors
  for word representation}.
\newblock In \emph{Proceedings of the 2014 Conference on Empirical Methods in
  Natural Language Processing ({EMNLP})}, pages 1532--1543, Doha, Qatar.
  Association for Computational Linguistics.

\bibitem[{Rubin and Berant(2021)}]{DBLP:journals/corr/abs-2010-12412}
Ohad Rubin and Jonathan Berant. 2021.
\newblock \href {https://doi.org/10.18653/v1/2021.naacl-main.29} {{S}m{B}o{P}:
  Semi-autoregressive bottom-up semantic parsing}.
\newblock In \emph{Proceedings of the 2021 Conference of the North American
  Chapter of the Association for Computational Linguistics: Human Language
  Technologies}, pages 311--324, Online. Association for Computational
  Linguistics.

\bibitem[{Shaw et~al.(2021)Shaw, Chang, Pasupat, and Toutanova}]{Shaw2020}
Peter Shaw, Ming-Wei Chang, Panupong Pasupat, and Kristina Toutanova. 2021.
\newblock \href {https://doi.org/10.18653/v1/2021.acl-long.75} {Compositional
  generalization and natural language variation: Can a semantic parsing
  approach handle both?}
\newblock In \emph{Proceedings of the 59th Annual Meeting of the Association
  for Computational Linguistics and the 11th International Joint Conference on
  Natural Language Processing (Volume 1: Long Papers)}, pages 922--938, Online.
  Association for Computational Linguistics.

\bibitem[{Shi et~al.(2020)Shi, Ng, Wang, Zhu, Li, Wang, dos Santos, and
  Xiang}]{DBLP:journals/corr/abs-2012-10309}
Peng Shi, Patrick Ng, Zhiguo Wang, Henghui Zhu, Alexander~Hanbo Li, Jun Wang,
  C{\'{\i}}cero~Nogueira dos Santos, and Bing Xiang. 2020.
\newblock \href {http://arxiv.org/abs/2012.10309} {Learning contextual
  representations for semantic parsing with generation-augmented pre-training}.
\newblock \emph{CoRR}, abs/2012.10309.

\bibitem[{Speer and Havasi(2012)}]{speer-havasi-2012-representing}
Robyn Speer and Catherine Havasi. 2012.
\newblock \href
  {http://www.lrec-conf.org/proceedings/lrec2012/pdf/1072_Paper.pdf}
  {{Representing General Relational Knowledge in {C}oncept{N}et 5}}.
\newblock In \emph{Proceedings of the Eighth International Conference on
  Language Resources and Evaluation ({LREC}'12)}, pages 3679--3686, Istanbul,
  Turkey. European Language Resources Association (ELRA).

\bibitem[{Suhr et~al.(2020)Suhr, Chang, Shaw, and Lee}]{Suhr2020}
Alane Suhr, Ming-Wei Chang, Peter Shaw, and Kenton Lee. 2020.
\newblock \href {https://doi.org/10.18653/v1/2020.acl-main.742} {Exploring
  unexplored generalization challenges for cross-database semantic parsing}.
\newblock In \emph{Proceedings of the 58th Annual Meeting of the Association
  for Computational Linguistics}, pages 8372--8388, Online. Association for
  Computational Linguistics.

\bibitem[{Wang et~al.(2020)Wang, Shin, Liu, Polozov, and Richardson}]{Wang2019}
Bailin Wang, Richard Shin, Xiaodong Liu, Oleksandr Polozov, and Matthew
  Richardson. 2020.
\newblock \href {https://doi.org/10.18653/v1/2020.acl-main.677} {{{RAT-SQL}:
  Relation-Aware Schema Encoding and Linking for Text-to-{SQL} Parsers}}.
\newblock In \emph{Proceedings of the 58th Annual Meeting of the Association
  for Computational Linguistics}, pages 7567--7578, Online. Association for
  Computational Linguistics.

\bibitem[{Yu et~al.(2018)Yu, Zhang, Yang, Yasunaga, Wang, Li, Ma, Li, Yao,
  Roman, Zhang, and Radev}]{Yu2018a}
Tao Yu, Rui Zhang, Kai Yang, Michihiro Yasunaga, Dongxu Wang, Zifan Li, James
  Ma, Irene Li, Qingning Yao, Shanelle Roman, Zilin Zhang, and Dragomir Radev.
  2018.
\newblock \href {https://doi.org/10.18653/v1/D18-1425} {{S}pider: A large-scale
  human-labeled dataset for complex and cross-domain semantic parsing and
  text-to-{SQL} task}.
\newblock In \emph{Proceedings of the 2018 Conference on Empirical Methods in
  Natural Language Processing}, pages 3911--3921, Brussels, Belgium.
  Association for Computational Linguistics.

\bibitem[{Zelle and Mooney(1996)}]{data-geography-original}
John~M Zelle and Raymond~J Mooney. 1996.
\newblock \href {http://dl.acm.org/citation.cfm?id=1864519.1864543} {{Learning
  to Parse Database Queries Using Inductive Logic Programming}}.
\newblock In \emph{Proceedings of the Thirteenth National Conference on
  Artificial Intelligence - Volume 2}, pages 1050--1055.

\bibitem[{Zhao et~al.(2021)Zhao, Cao, and
  Zhao}]{DBLP:journals/corr/abs-2101-09901}
Liang Zhao, Hexin Cao, and Yunsong Zhao. 2021.
\newblock \href {http://arxiv.org/abs/2101.09901} {{GP:} context-free grammar
  pre-training for text-to-sql parsers}.
\newblock \emph{CoRR}, abs/2101.09901.

\bibitem[{Zhong et~al.(2017)Zhong, Xiong, and Socher}]{zhongSeq2SQL2017}
Victor Zhong, Caiming Xiong, and Richard Socher. 2017.
\newblock {Seq2SQL: Generating Structured Queries from Natural Language using
  Reinforcement Learning}.
\newblock \emph{CoRR}, abs/1709.0.

\end{thebibliography}
\bibliographystyle{acl_natbib}

\appendix



\end{document}